\documentclass[runningheads]{llncs}
\usepackage[T1]{fontenc}
\usepackage{graphicx,verbatim}
\usepackage{amsmath}      % \text, \lVert, \rVert (llncs does not load amsmath)
\usepackage{xcolor}       % \color{blue}, \textcolor{blue}, gray!8
\usepackage{colortbl}     % \columncolor (column type M)
\usepackage{array}        % m{} columns, \arraybackslash
\usepackage{booktabs}     % \toprule, \midrule, \bottomrule
\usepackage{multirow}     % \multirow (Table 1)
\usepackage{siunitx}      % S columns (Table 1)
\usepackage{placeins}     % \FloatBarrier

\newlength{\origtextwidth}
\newcolumntype{C}{>{\centering\arraybackslash}m{0.173\origtextwidth}}                 % was p{...}, now m{...}
\newcolumntype{M}{>{\centering\arraybackslash\columncolor{gray!8}}m{0.085\origtextwidth}}
\newcommand{\RegRow}[8]{%
  \begin{tabular}[c]{@{}c@{}}
    \textbf{Seq.\,#1}\\[3pt]
    {\footnotesize pair #2}
  \end{tabular}
  & \includegraphics[width=\linewidth]{#3}
  & \includegraphics[width=\linewidth]{#4}
  & \includegraphics[width=\linewidth]{#5}
  & \includegraphics[width=\linewidth]{#6}
  & \includegraphics[width=\linewidth]{#7}
  & \includegraphics[width=\linewidth]{#8}
  \\[8pt]
}

\begin{document}
\title{Evaluating Transformation Models for pCLE Mosaic Registration}
\begin{comment}  %% Removed for anonymized MICCAI submission
\author{First Author\inst{1}\orcidID{0000-1111-2222-3333} \and
Second Author\inst{2,3}\orcidID{1111-2222-3333-4444} \and
Third Author\inst{3}\orcidID{2222--3333-4444-5555}}
%
\authorrunning{F. Author et al.}
\institute{Princeton University, Princeton NJ 08544, USA \and
Springer Heidelberg, Tiergartenstr. 17, 69121 Heidelberg, Germany
\email{lncs@springer.com}\\
\url{http://www.springer.com/gp/computer-science/lncs} \and
ABC Institute, Rupert-Karls-University Heidelberg, Heidelberg, Germany\\
\email{\{abc,lncs\}@uni-heidelberg.de}}

\end{comment}

% ---- Anonymized block for double-blind review: swap the comments between this block and the block below before submitting ----
%\author{Anonymized Authors}  %% Added for anonymized MICCAI submission
%\authorrunning{Anonymized Author et al.}
%\institute{Anonymized Affiliations \\
%    \email{email@anonymized.com}}
%
% ---- Author block with names and affiliations (NOT anonymous) ----
\author{Ahmed Aboelela\inst{1} \and
Johannes Barcsay\inst{2} \and
Jana Friedhoff\inst{3} \and
Miguel Gonçalves\inst{3} \and
Alexander Hann\inst{2} \and
Katharina Breininger\inst{1}}
\authorrunning{A. Aboelela et al.}
\institute{Center for Artificial Intelligence and Data Science (CAIDAS), Julius-Maximilians-Universität Würzburg, Würzburg, Germany \and
Department of Internal Medicine II, Interventional and Experimental Endoscopy (InExEn), University Hospital Würzburg, Würzburg, Germany \and
Department of Otorhinolaryngology, Plastic and Aesthetic Operations, University Hospital Würzburg, Würzburg, Germany}
% TODO (not in the original paper): corresponding-author \email{...} and \orcidID{...} per author
  
\maketitle              % typeset the header of the contribution
\begin{abstract}
Confocal Laser Endomicroscopy (CLE) provides real-time,
cellular-resolution optical biopsy but has a narrow field of view, which
image mosaicing can extend to provide anatomical context. Because of
line-by-line acquisition, probe motion, and probe--tissue interaction,
frame alignment generally requires a non-linear transformation whose
accuracy is difficult to quantify: flexible transformation models can fit
intensity features and noise, so appearance-based metrics such as
Normalized Cross-Correlation (NCC) can improve without a genuine gain in
geometric accuracy. We
therefore establish a dataset of 132 frame pairs across fourteen pCLE
sequences from 4 patients with manually annotated landmark
correspondences, so that Target Registration Error (TRE) can serve as a
geometrically grounded complement to NCC. We assess the effect of
progressively increasing the transformation model's degrees of freedom,
from translation to Thin Plate Spline (TPS), and of six feature-matching
backends spanning classical (Shi-Tomasi, Lucas-Kanade) and learned
(SuperPoint, SuperGlue, LightGlue, LoFTR, RoMa) approaches. Translation
and rigid models prove insufficient under tissue deformation, while TPS
with random sampling achieves the strongest landmark-derived alignment of
the evaluated configurations; among the learned matchers, used without
fine-tuning, only RoMa offers a robust, if modest, advantage over
other methods. At the sequence level, pairwise registration quality
proved an unreliable predictor of final mosaic quality, so mosaic quality must be
evaluated directly rather than inferred from pairwise metrics.

% Authors must provide keywords and are not allowed to remove this Keyword section.

\end{abstract}
\section{Introduction}
{
Cancer-related surgery in sensitive anatomical regions such as the larynx can degrade postoperative quality of life through the loss of healthy tissue, making accurate intraoperative identification of tumor boundaries critical. Frozen section, the gold standard for assessing resection margins, requires tissue removal, samples the margin only at discrete locations, and delays the result, increasing both operating room utilization and cost.

Confocal Laser Endomicroscopy (CLE) enables real time, in situ, in vivo optical imaging at cellular resolution, but its small field of view requires considerable mental effort from the surgeon to relate the current image to the surrounding anatomy. Three properties of the imaging process further complicate the alignment of successive frames. The probe must be held in contact with the tissue under pressure, which deforms the tissue; image intensity varies between frames with applied pressure and with the concentration of administered fluorescein, which stains the extracellular space so that cells appear dark against a bright background; and the signal is acquired line by line, so probe motion between successive scan lines can make parts of the tissue appear contracted or stretched~\cite{aubreville2019,vercauteren:inria-00616123}.
}

To increase the spatial context available for interpretation, successive pCLE frames can be stitched into a mosaic. Evaluating mosaic quality is difficult because deformation varies across a sequence and noise is pervasive: appearance-based metrics such as Normalized Cross-Correlation (NCC) may improve as the transformation model gains degrees of freedom without the mosaic improving since the added freedom may simply fit noise or align repeating but mismatched cell patterns. Manually matched landmarks between frame pairs provide a geometric alternative: cell junctions remain identifiable under moderate deformation and noise, and the resulting Target Registration Error (TRE) measures geometric consistency independently of appearance differences.

In this paper, we create a dataset with matching landmarks across pairs of pCLE video frames to assess two aspects: First, we investigate the effect of progressively increasing the degrees of freedom of the transformation model, from translation-only through to Thin Plate Spline on NCC and TRE; second, we evaluate the use of automated feature extraction and matching approaches, moving from classical methods such as Shi-Tomasi \cite{shi-tomasi-features} and Lucas-Kanade \cite{Lucas-kanade} to modern approaches such as SuperPoint \cite{superpoint} and LoFTR \cite{loftr}. We further evaluate sequential mosaicing without global optimization to assess execution reliability, geometric and photometric consistency, and their relationship to pairwise registration performance. We will release our code open source, as, to the best of our knowledge, no existing open-source library for CLE mosaicing addresses both nonrigidity and multiple transformation models.

\section{Related Work}
Vercauteren et al.~\cite{Vercauteren2006Robust} and Loewke et al.~\cite{loewkeInVivo} established the rigid-plus-nonrigid decomposition underlying most subsequent pCLE mosaicing: global alignment (Lie-group estimation or soft-constraint least-squares) followed by a local deformation correction (demons or radial-basis-function surfaces). Gong et al.~\cite{Gong2021RobustMosaicing} and Gong et al.~\cite{Gong2022IntensityBased} instead target the similarity metric driving this correction, replacing intensity-difference objectives with context- and texture-weighted variants (CWCR, TESCV) to improve robustness under intensity fluctuation and variable contact force. Most recently, Hao et al.~\cite{Hao:26} identify the same transformation-model inadequacy motivating our DOF hierarchy, observing that affine models capture only global average deformation while grid-based free-form deformation leaves fine local variation uncorrected, and propose a feature-based non-rigid method that rejects mismatches using a motion-consistency constraint derived from the pCLE point-scanning mechanism. Benchmarked against Demons, CWCR, TESCV, and several feature matchers (GMS, MAGSAC++, AdaLAM, LightGlue) using both structural similarity and manually annotated landmark distance, their best pooled performer, SIFT combined with LightGlue, still shows visible local misalignment on qualitative inspection, directly supporting the concern that photometric or structural similarity can be satisfied without correct underlying geometry. Complementary work addresses adjacent aspects of the same pipeline without engaging the transformation-model question directly: Rosa et al.~\cite{rosa} and Zhang et al.~\cite{Zhang} robustify the translation-only online topology inference by fusing image-based registration with an independent robot or camera motion estimate in a Kalman filter, validated primarily on rigid or phantom targets against external tracking, optical microscopy, or grid-pattern ground truth, with deformable ex vivo tissue used only for qualitative demonstration; Mahé et al.~\cite{mahe:hal-01208437} treat spatial alignment as a fixed prerequisite and instead preserve dynamic content such as capillary blood flow during mosaic composition, evaluated by clinician Likert-scale rating rather than geometric ground truth.

% In each case, however, transformation complexity and similarity metric are varied together rather than independently. With the exception of~\cite{Hao:26}, who validate against manually annotated landmarks, accuracy is reported using the same photometric criterion optimized during registration, leaving open whether reported gains reflect added degrees of freedom or metric saturation; and even~\cite{Hao:26}, despite this independent ground truth, still vary registration paradigm and correspondence backend together rather than holding them fixed across a controlled DOF hierarchy. Furthermore, while \cite{Hao:26} report manual reference point annotation for error assessment and \cite{mahe:hal-01208437} describe the comparison with a manually aligned ``oracle'' sequence, there is little further information about coverage or annotation strategy available. Additionally, evaluation of correct probe motion (not tissue motion) was mostly performed under controlled laboratory conditions and/or phantoms, whether via a robot-controlled trajectory~\cite{rosa,Zhang} or a manual translation stage~\cite{Hao:26}, and may not fully reflect tissue deformation during clinical use of pCLE.

In each case, however, transformation complexity and matching backend are varied together rather than independently. With the exception
of~\cite{Hao:26}, accuracy is reported using the same photometric
criterion optimized during registration, leaving open whether reported
gains reflect added degrees of freedom or improved feature extraction and matching backend; even
\cite{Hao:26}, despite validating against manual landmarks, vary
registration paradigm and correspondence backend together rather than
holding them fixed across a controlled DOF hierarchy, and neither their
annotations nor the manually aligned oracle sequence of~\cite{mahe:hal-01208437}
come with information on coverage or annotation strategy. Finally, these
evaluations mostly assess probe motion rather than tissue deformation,
under laboratory conditions or on phantoms via a robot-controlled
trajectory~\cite{rosa,Zhang} or a manual translation
stage~\cite{Hao:26}, and may not reflect clinical use of pCLE.

% Vercauteren et al.~\cite{Vercauteren2006Robust} and Loewke et al.~\cite{loewkeInVivo} established the rigid-plus-nonrigid decomposition underlying most subsequent pCLE mosaicing: global alignment (Lie-group estimation or soft-constraint least-squares) followed by a local deformation correction (demons or radial-basis-function surfaces). 
% Gong et al.~\cite{Gong2021RobustMosaicing} and Gong et al.~\cite{Gong2022IntensityBased} instead target the similarity metric driving this correction, replacing intensity-difference objectives with context- and texture-weighted variants (CWCR, TESCV) to improve robustness under intensity fluctuation and variable contact force. 
% In each case, however, transformation complexity and similarity metric are varied together rather than independently, and
% accuracy is reported using the same photometric criterion optimized during registration, leaving open whether reported gains reflect added degrees of freedom or metric saturation. Additionally, evaluation of correct probe motion (not tissue motion) was performed under controlled laboratory conditions and/or phantoms and may not fully reflect tissue deformation during clinical use of pCLE.

Feature extraction and matching methods can be categorized by how
densely they populate the image, a property directly relevant to pCLE's
low-texture, noisy field of view. Sparse detector-based
pipelines, from Shi-Tomasi corners tracked by Lucas-Kanade optical
flow~\cite{shi-tomasi-features,Lucas-kanade} to the learned SuperPoint
detector~\cite{superpoint} with the attentional matchers
SuperGlue~\cite{superglue} and LightGlue~\cite{lightglue} all
assume that a detector fires independently on the same physical
locations in both views, an assumption that breaks down in
low-texture regions. Detector-free methods avoid this by matching dense
features directly, on a coarse grid in LoFTR~\cite{loftr} and at full
pixel density with pretrained foundation-model features in
RoMa~\cite{roma}. To our knowledge, none has been evaluated on
CLE or a comparably low-texture,
cellular-resolution modality, leaving open whether their robustness transfers to pCLE.

\section{Methods}

\subsection{Dataset}

\begin{figure}[t]
  \begin{minipage}{1.0\linewidth }
  \centering
    \includegraphics[width=0.9\linewidth]{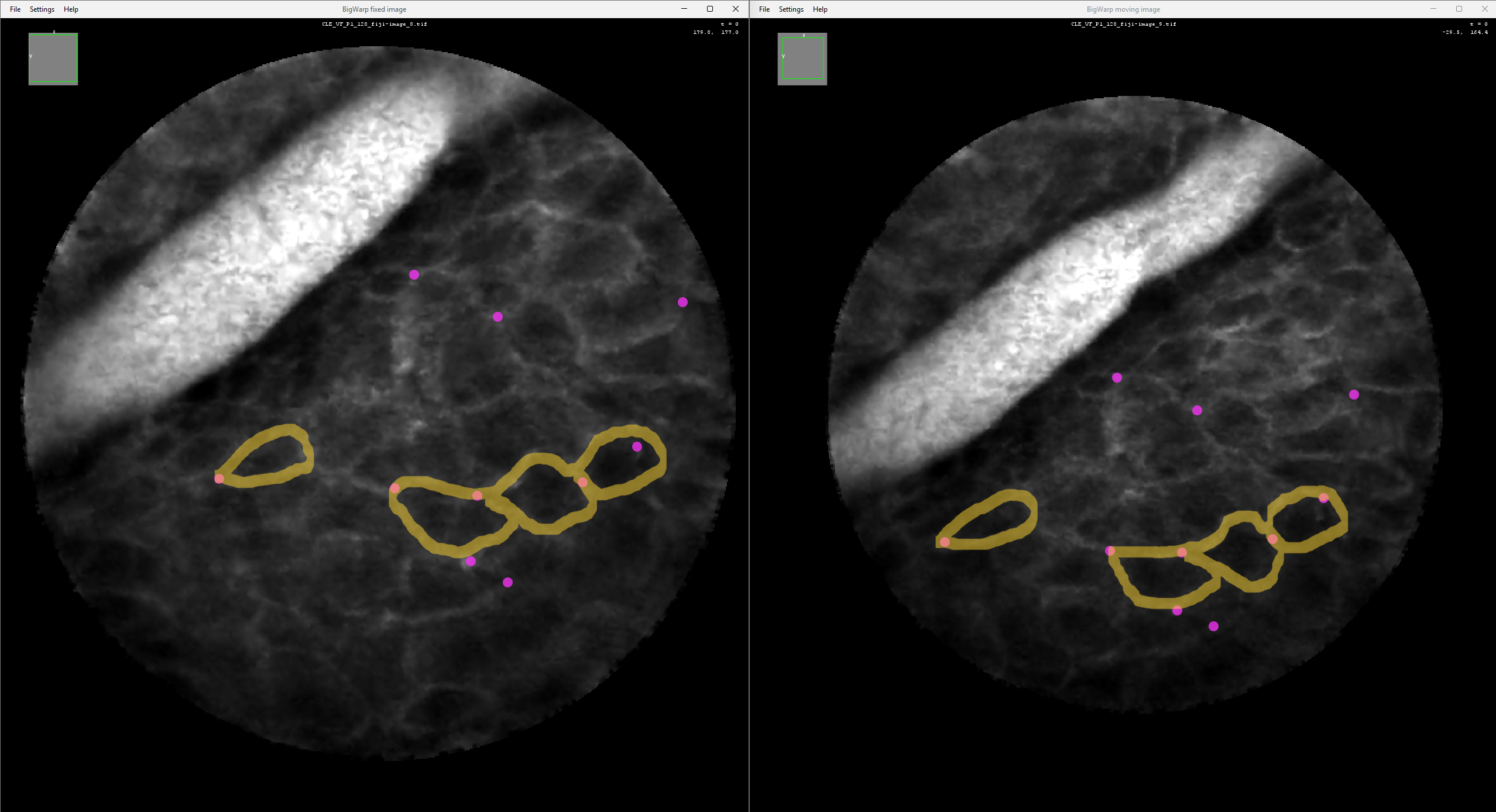}
  \end{minipage}\hfill
  \caption{Sequence 128, frames 8–9, with candidate anchor structures marked. Distinctive corner structures common to both frames are identified as starting points for placement.}
  \label{fig:pCLE_pair_eye_from_CLE_VF_P1_128_frames_8_9_with_marking}
\end{figure}

\begin{figure}[t]
  \centering
  \begin{minipage}{1.0\linewidth }
  \centering
    \includegraphics[width=0.9\linewidth]{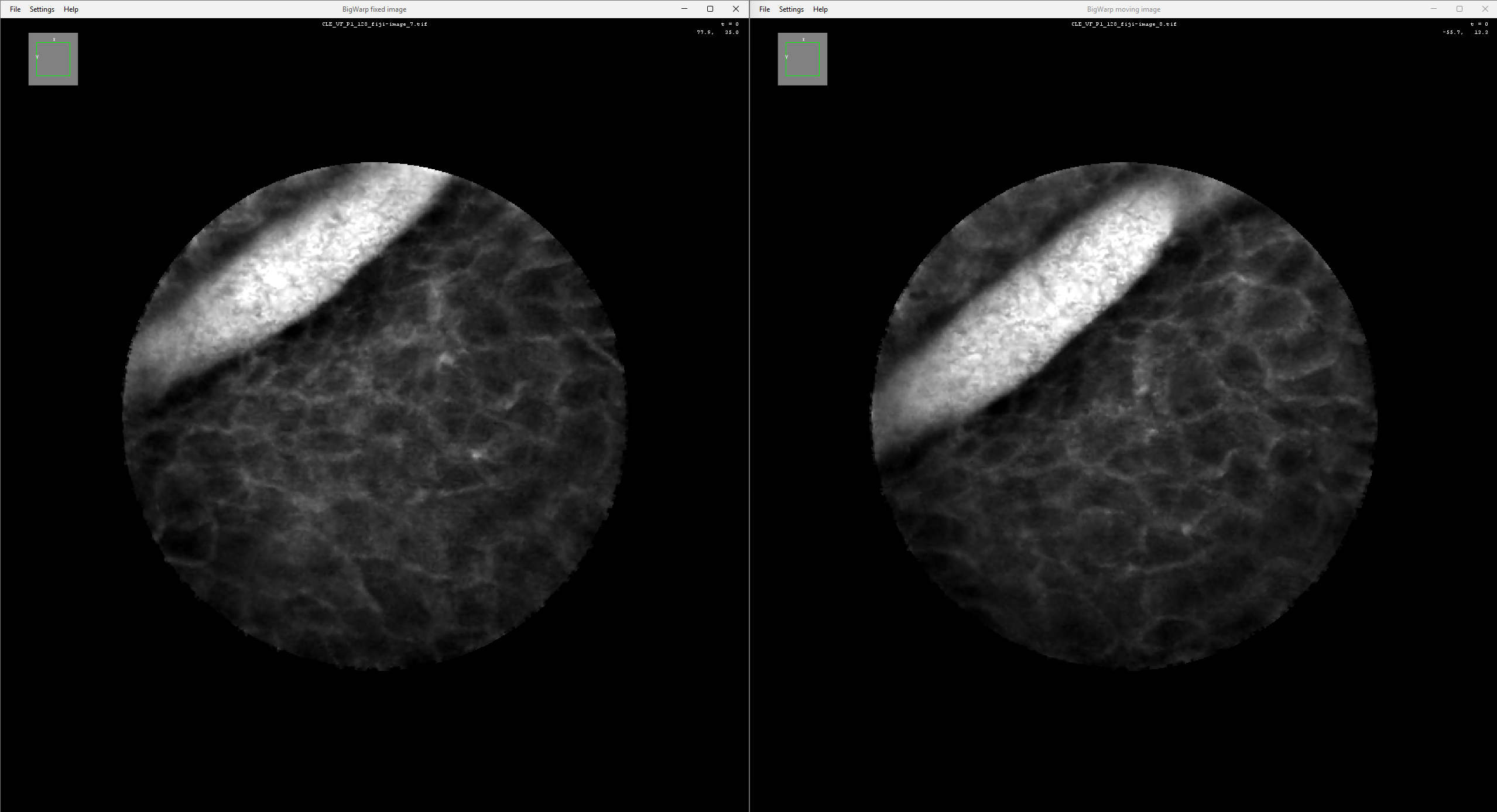}
  \end{minipage}\hfill
  \caption{Frame pair from Sequence 128. The deformation/motion blur between the two frames is severe enough that reliable matching keypoints cannot be identified by eye, illustrating a case excluded from annotation.}
  \label{fig:pCLE_pair_that_cant_be_registered_by_eye_from_CLE_VF_P1_128_frames_7_8}
\end{figure}

The dataset comprises 132 annotated frame pairs from fourteen pCLE sequences acquired from four patients with suspected or confirmed laryngeal cancer at Universitätsklinikum Erlangen using a Cellvizio system (Mauna Kea Technologies, Paris, France). Table \ref{tab:pairs_per_sequence_basic_summary} summarizes patient–sequence assignments and landmark counts. Correspondences were manually annotated using BigWarp \cite{bigwarp} by identifying distinctive cell-corner structures and tracing them to their counterparts in the paired frame (Figure \ref{fig:pCLE_pair_eye_from_CLE_VF_P1_128_frames_8_9_with_marking}). Corners provided identifiable anchors despite noisy cell boundaries, although changes in their appearance introduced localization uncertainty. Pairs with deformation or image degradation severe enough to prevent reliable visual matching were excluded (Figure \ref{fig:pCLE_pair_that_cant_be_registered_by_eye_from_CLE_VF_P1_128_frames_7_8}), biasing the evaluation toward annotatable tissue regions and frame pairs. Probe fouling and abrupt relocation were also excluded. The dataset therefore contains selected pairs rather than complete sequences and does not cover the full range of clinical acquisition conditions. Patient demographics were unavailable because the data were anonymized, and tissue categories (benign, precancerous, or cancerous) were not verified because annotation was performed by non-physicians. No ground-truth probe trajectory or reference mosaic was available, limiting sequence-level evaluation to internal consistency rather than absolute pose or mosaic accuracy.


\begin{table}[!t]
\centering
\caption{Per-sequence summary of annotated frame pairs and the number of manually annotated landmarks per pair. Mean, median, minimum, and maximum are computed across all registered pairs within each sequence.\label{tab:pairs_per_sequence_basic_summary}}
\footnotesize
\setlength{\tabcolsep}{2pt}
\begin{tabular}{c c S[table-format=3.0] S[table-format=2.0] S[table-format=2.0] S[table-format=2.0] S[table-format=2.0]}
\toprule
\textbf{Patient} & \textbf{Sequence} & \textbf{\# Pairs} & \textbf{Mean} & \textbf{Median} & \textbf{Min} & \textbf{Max} \\
\midrule
\multirow{6}{*}{P1 ($6$)}
 & CLE\_VF\_P1\_122 & 14  & 10.07 & 10.0 & 8  & 12 \\
 & CLE\_VF\_P1\_128 & 18  & 10.39 & 10.0 & 8  & 16 \\
 & CLE\_VF\_P1\_137 & 6   & 11.17 & 11.0 & 10 & 12 \\
 & CLE\_VF\_P1\_172 & 5   & 11.20 & 11.0 & 9  & 13 \\
 & CLE\_VF\_P1\_182 & 10  & 14.60 & 14.5 & 11 & 21 \\
 & CLE\_VF\_P1\_185 & 6   & 10.67 & 10.5 & 8  & 14 \\
\midrule
\multirow{3}{*}{P2 ($3$)}
 & CLE\_VF\_P2\_123 & 8   & 10.50 & 10.5 & 9  & 12 \\
 & CLE\_VF\_P2\_129 & 8   & 11.13 & 11.0 & 10 & 14 \\
 & CLE\_VF\_P2\_134 & 9   & 10.11 & 10.0 & 9  & 11 \\
\midrule
\multirow{3}{*}{P3 ($3$)}
 & CLE\_VF\_P3\_119 & 8   & 10.13 & 10.0 & 8  & 12 \\
 & CLE\_VF\_P3\_124 & 8   & 10.25 & 10.0 & 10 & 11 \\
 & CLE\_VF\_P3\_130 & 8   & 10.25 & 10.0 & 9  & 12 \\
\midrule
\multirow{2}{*}{P4 ($2$)}
 & CLE\_VF\_P4\_120 & 8   & 9.50  & 10.0 & 7  & 11 \\
 & CLE\_VF\_P4\_125 & 16  & 10.44 & 10.0 & 10 & 12 \\
\bottomrule
\end{tabular}
\end{table}
% {\color{blue}
% We construct a dataset with 132 annotated image pairs from the fourteen sequences (Table~\ref{tab:pairs_per_sequence_basic_summary}); the landmarks allow us to compare TRE and NCC for every transformation configuration, revealing whether an apparent gain in appearance is accompanied by a gain in geometric accuracy.
% }

% Potentially add back: approaches to assess dataset quality.

\subsection{Transformation Models}\label{sec:methods-transformation-models}

When registering pCLE sequences automatically, image similarity metrics like NCC can saturate for a given transformation model without reaching sufficient anatomical alignment. The plateau may reflect the model exhausting its capacity to represent the true motion, or it may reflect a limit on the quality of the correspondences supplied to it, regardless of the model's flexibility. These two explanations call for different remedies: increasing the degrees of freedom in the former case, or improving the feature correspondences in the latter. The saturated NCC value alone does not indicate which explanation applies.

To disambiguate these two factors, we first establish, using the available landmark annotations, the accuracy achievable at each level of model complexity. Each frame pair is registered under four transformation models with increasing flexibility: translation ($2$ DOF); rigid registration, which combines translation with rotation ($3$ DOF); affine ($6$ DOF); and a Thin Plate Spline (TPS) model, whose degrees of freedom scale with the number of correspondences rather than being fixed. The hierarchy also has a diagnostic use: because a rigid model cannot represent local tissue deformation while a non-rigid model can, a substantially higher TRE or lower NCC under the rigid model than under a
non-rigid one typically indicates significant deformation in the pair. To characterize the alignment performance attainable with each transformation model when errors from automated correspondence estimation are excluded, the transformation parameters were estimated from the manually annotated landmark correspondences. The resulting alignment provides a landmark-derived reference indicating the performance that each model can plausibly achieve under the available annotations and estimation procedure, rather than an absolute upper bound. Because only one set of annotations was available, inter-annotator agreement and intra-annotator repeatability could not be quantified. The reference may therefore remain affected by annotation uncertainty and should primarily be used to contextualize the results obtained with automated feature-matching backends.

For the translation, rigid, and affine models, the transformation is
overdetermined by the annotated landmarks and is estimated with
RANSAC~\cite{ransac}, which retains the minimal-sample hypothesis
with the most inliers under a fixed reprojection threshold and therefore
treats some landmarks as outliers. Since TPS can overfit the landmarks,
it is fitted in three ways: to the inliers of an affine RANSAC estimate
(TPS Affine+RANSAC), to a random subset of the landmarks (TPS random
sampling), or to all landmarks directly (TPS raw).

% For the translation, rigid, and affine models, the transformation is overdetermined by the number of annotated landmarks. We therefore use RANSAC \cite{ransac}: the minimum number of landmark pairs needed to determine a given model's parameters is sampled, the resulting transformation is scored by how many of the remaining landmarks it reprojects within a fixed distance threshold, and the transformation with the most inliers is kept. Once RANSAC is applied to obtain the transformation from landmarks, this procedure will inevitably treat some landmarks as outliers. Since TPS may overfit given the landmarks, we apply three strategies. First, inlier points can be extracted through a combination of affine estimation and RANSAC, after which the TPS is fitted to these inliers. Second, points can be extracted through random sampling, after which the TPS is fitted to these randomly sampled points. Third, the TPS can be fitted to all landmarks directly.

\subsection{Feature Extraction \& Matching} \label{sec:methods-matchers}
We evaluate six correspondence backends, all off-the-shelf without
fine-tuning: Shi-Tomasi~\cite{shi-tomasi-features} and SuperPoint~\cite{superpoint}
keypoints tracked with Lucas-Kanade~\cite{Lucas-kanade}, SuperPoint
keypoints matched with SuperGlue~\cite{superglue} and
LightGlue~\cite{lightglue}, and the detector-free
LoFTR~\cite{loftr} and RoMa~\cite{roma}.

Each of the six backends is evaluated under all four transformation models using TRE and NCC, isolating which methods underperform and which perform best. All evaluation here is pairwise, without global optimization.

For automated correspondences, RANSAC is applied analogously: the minimal point sample each model requires is drawn, and the hypothesis with the highest inlier count is retained. TPS again uses the affine-RANSAC-inlier strategy described above; however, because affine's limited degrees of freedom can misclassify genuine nonrigid inliers as outliers, we also rely on TPS (Random Sampling): correspondences are drawn randomly and used to fit a TPS transformation, repeated $n$ times, with the transformation yielding the highest NCC retained as the final result.

% Correspondences are filtered identically across models, analogous to the selection for manually selected landmarks: RANSAC draws the minimal point sample each model requires and keeps the hypothesis with the highest inlier count. TPS cannot be fit via RANSAC directly, so inliers are instead obtained from an affine RANSAC fit and then passed to TPS. This risks excluding points that are genuine inliers under a nonrigid model but appeared as outliers only because affine's limited degrees of freedom could not explain them. To guard against this, we introduce TPS (Random Sampling): a random percentage of correspondences is drawn and used to fit a TPS transformation, and this process is repeated $n$ times, with the transformation yielding the highest normalized cross-correlation (NCC) retained as the final result.

\subsection{Sequence-Level Mosaicing}

To assess how far pairwise registration alone can support pCLE
mosaicing, complete sequences were reconstructed without global
optimization. Each sequence was represented as a directed, non-branching
chain; the first frame defined the coordinate system, and subsequent
pairwise transformations were accumulated as
\begin{equation}
T_{i+1} = T_i \circ R_{i+1 \rightarrow i},
\label{eq:accumulation}
\end{equation}
where $R_{i+1 \rightarrow i}$ maps the moving frame to its predecessor.
Translation, rigid, and affine transformations were composed exactly;
To reduce computational cost and keep evaluation time manageable, accumulated TPS deformations were approximated on a dense interpolation grid, introducing interpolation error relative to direct evaluation of the composed warps. Frames were resampled within their circular pCLE
support and blended by normalized weighted accumulation. A pairwise
registration failure terminated the current segment and started a new
independently anchored one, preserving all successfully processed frames
while exposing failure-induced fragmentation.

\subsection{Evaluation Protocol}\label{sec:methods-evaluation-protocol}
\subsubsection{Dataset Quality Assurance}

As described previously, landmarks are placed on corners that can be
re-identified in the subsequent frame. This selection may underrepresent
regions with lower image quality or stronger deformation, and since
deformation varies across the field of view, landmarks concentrated in a
small region constrain a deformable transformation less effectively than
fewer landmarks spread across the image. We therefore characterize the
spatial distribution of each annotated point set with three complementary
statistics, all computed over the circular field-of-view mask $\Omega$
rather than the square image frame: the convex hull coverage ratio
$C_{\text{hull}}$, the fraction of $\Omega$ spanned by the landmarks; the
Clark--Evans index $R$, the ratio of the observed mean nearest-neighbor
distance to its edge-corrected expectation under complete spatial
randomness~\cite{Donnelly1978}, where $R < 1$ indicates landmarks more
clustered than chance and $R > 1$ a more dispersed pattern; and the
coverage radius $d_{\max}$, the largest distance from any point in
$\Omega$ to its nearest landmark, which bounds how far a tissue location
can lie from a ground-truth constraint. Full definitions and the edge
correction are given in the Suppl. Mat., Sec.~1.1.

\subsubsection{Quantification of Alignment}
Two metrics are used to evaluate the methods introduced in Sections~\ref{sec:methods-transformation-models} and \ref{sec:methods-matchers}: Target Registration Error (TRE) and Normalized Cross-Correlation (NCC). For a landmark $\mathbf{x}_i$ annotated in the source frame and its independently annotated correspondence $\mathbf{y}_i$ in the target frame, TRE under an estimated transformation $T$ is

\[
\mathrm{TRE}_i = \lVert T(\mathbf{x}_i) - \mathbf{y}_i \rVert_2.
\]

% Because $\mathbf{y}_i$ is annotated by direct inspection of the target frame rather than derived from $T$, it serves as an independent baseline against which the accuracy of $T(\mathbf{x}_i)$ is scored, rather than a target that would shift with the transformation being evaluated. %For the transformations based on manually annotated landmarks, TRE 

NCC, by contrast, is computed directly from the photometric content of the two frames being aligned,

\[
\mathrm{NCC}(I_1, I_2) = \frac{\sum(I_1-\bar I_1)(I_2-\bar I_2)}{\sqrt{\sum(I_1-\bar I_1)^2\sum(I_2-\bar I_2)^2}},
\]

evaluated over the valid overlap region between the warped source and target frames.

Pairwise geometric accuracy was evaluated using three repetitions of leave-one-out cross-validation. In each fold, one landmark was excluded, the transformation was fitted to the remaining landmarks, and the excluded landmark’s Euclidean reprojection error was measured. Automatic methods were evaluated on the same held-out landmark, although manual landmarks were not used for fitting. Each landmark was therefore evaluated three times, and its median error was retained. The pair-level error was the mean of these median errors across all landmarks.
NCC was evaluated separately on the transformation fitted with all
landmarks, or on the full automatic registration, which does not affect
the cross-validated landmark error since NCC measures photometric
agreement rather than held-out landmark accuracy.

% \subsubsection{Diagnostic Interpretation of the Transformation Hierarchy}
% \label{sec:diagnostic-interpretation}

% Both NCC and TRE can indicate the degree of tissue deformation by comparing the registration error obtained under a rigid transformation model to that obtained under a non-rigid model. Because a rigid model cannot account for local tissue deformation while a non-rigid model can, a substantially higher error under the rigid model relative to the non-rigid model typically indicates significant deformation in the image.

\subsubsection{Correlation Analysis}\label{sec:correlation_analysis}
To assess the correlation between geometric and photometric registration quality, we compute Pearson's $r$ and Spearman's rank $\rho_s$ between per-pair TRE and $1-\text{NCC}$ within each sequence.

\subsubsection{Statistical and Multi-Objective Performance Comparison}

Methods are compared separately for TRE and NCC using paired, sequence-level non-parametric analyses. Each method's performance is summarized per sequence by the mean over the common set of annotated frame pairs, so that every sequence contributes one observation irrespective of its number of pairs. Within each fixed experimental setting, overall differences are assessed with the Friedman test~\cite{friedman}, with Kendall's $W$~\cite{kendall} as an omnibus effect size; if significant, all pairwise differences are tested with two-sided Wilcoxon signed-rank tests~\cite{wilcoxon}, Holm-adjusted~\cite{holm} across all pairwise comparisons for that metric and setting, and interpreted together with the direction and magnitude of the sequence-level differences.

TRE and NCC are additionally considered jointly through a sequence-level Pareto analysis with both objectives expressed as costs (TRE and $1-\text{NCC}$): within each sequence and setting, a method is non-dominated if no alternative is at least as good in both objectives and strictly better in one. The Pareto-front rate, the share of sequences on which a method is non-dominated, is a descriptive measure of how frequently it provides a competitive TRE--NCC trade-off; full definitions are in Suppl. Mat., Sec.~2.1.

\subsubsection{Sequence-Level Mosaic Evaluation}
\label{subsubsection:sequence_level_mosaic_evaluation}

Because no globally registered reference mosaic or ground-truth probe
trajectory was available, final mosaics were evaluated using three
measures of internal consistency (definitions in Suppl. Mat.,
Sec.~2.2). Leave-one-frame-out NCC (LOFO-NCC) compares each warped frame
with a reference mosaic built exclusively from the remaining frames.
Mosaic-coordinate landmark TRE measures the disagreement between
corresponding landmarks after projection into the shared mosaic
coordinate system, using leave-one-landmark-out refitting for
landmark-guided registration. Directional residual accumulation (DRA)
sums the mean signed landmark residuals of consecutive registration edges
within uninterrupted segments, summarizing coherent long-range drift.

For each transformation configuration, the association between the
sequence-level mean pairwise metric (pairwise NCC aggregated via a Fisher
$z$-transformed mean) and the corresponding mosaic-level metric was
quantified with Pearson and Spearman correlations across the 14
sequences. Confidence intervals used a cluster bootstrap over sequences
(2000 repetitions, 95\% percentile interval); $p$-values were
Benjamini--Hochberg adjusted ($\alpha = 0.05$) within each registration
source and transformation configuration, pooling across backends and the
three predictor--outcome relationships; robustness was assessed by
leave-one-sequence-out refitting; and configurations with fewer than 10
sequences were flagged as small-sample.

\section{Results \& Discussion}
We first characterize the annotated landmarks (Sec.~4.1) and the
landmark-derived reference performance of each transformation
configuration (Sec.~4.2), then compare the feature-matching backends
(Sec.~4.3), and finally evaluate mosaics generated without global
optimization (Sec.~4.4).
%Dataset Quality Evaluation sample 
% One interesting exception is the pair of registrations 18-19 \& 19-20 in sequence CLE\_VF\_P1\_122, where TPS (Raw) results in a very low NCC values, as per the analysis described in the supplementary material for the the landmarks distribution across the image. it shows that the pairs suffer from clustered landmark placement that compromised the resulting transformation. Figure \ref{fig:pCLE_pair_eye_from_CLE_VF_P1_122_frames_19_20_clustering_example} shows the landmarks being clustered in the center. The exact numbers are shown in Table 1 in the supplementary materials. Reading these metrics jointly helps flag frames that require additional annotation, since it is otherwise difficult to determine during annotation how many points a given frame needs: NCC never reaches 1 due to the presence of noise, and TRE cannot be used during annotation without introducing circularity. Because of this, the annotator's only real-time signal is visual inspection of transformation quality, which, as this example shows, can be misleading.

\subsection{Landmarks Distribution Evaluation}

Across the 14 sequences, convex hull coverage $C_{\text{hull}}$ averages between 20\% and 50\%, indicating that landmarks typically span less than half of the field of view, while the maximum empty-space distance $d_{\max}$ ranges between 66 and 125\,\textmu m; the Clark--Evans index $R$ falls mildly below 1 for about half of the sequences, indicating landmarks somewhat more clustered than a spatially random distribution would produce rather than severely aggregated, with Sequence P4-125 showing the best median dispersion. $C_{\text{hull}}$ and $d_{\max}$ are conservative in opposite directions, the former a lower bound on true coverage, the latter an upper bound on the true annotation gap, so in both cases the reported value can only make the annotations appear as well-covered as, or worse-covered than, the truth, never better. Per-sequence distributions of all three statistics, and their joint relationship across point sets, are reported in the Suppl. Material (Sec.~3.1, Fig.~1). These coverage statistics should be read alongside registration-quality indicators: a pair combining poor coverage (low $C_{\text{hull}}$, high $d_{\max}$) with low NCC and high TRE signals severe deformation and should be flagged for re-annotation.

\begin{figure}[!t]
\centering
\resizebox{\textwidth}{!}{
\setlength{\tabcolsep}{3pt}          % tighten cell padding to reclaim horizontal space for 6 columns
\begin{tabular}{M|CCCCCC}
 & \textbf{(a) Translation} & \textbf{(b) Rigid} & \textbf{(c) Affine} & \textbf{(d) TPS (Aff.+RANSAC)} & \textbf{(e) TPS (Raw)} & \textbf{(f) TPS (Random Sampling)} \\
\hline
\RegRow{128}{20--21}%
  {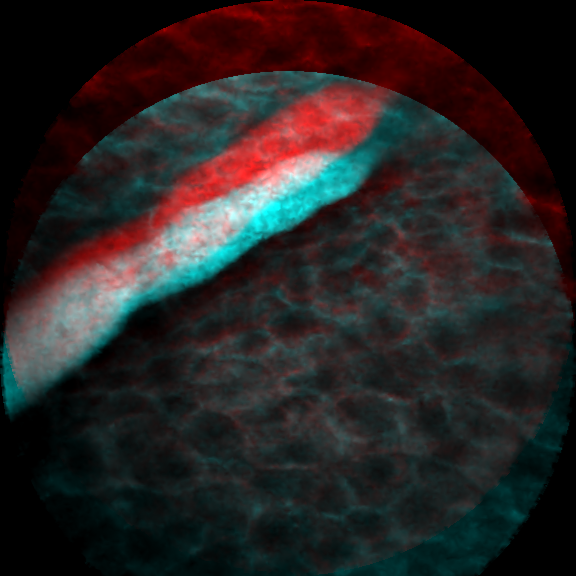}%
  {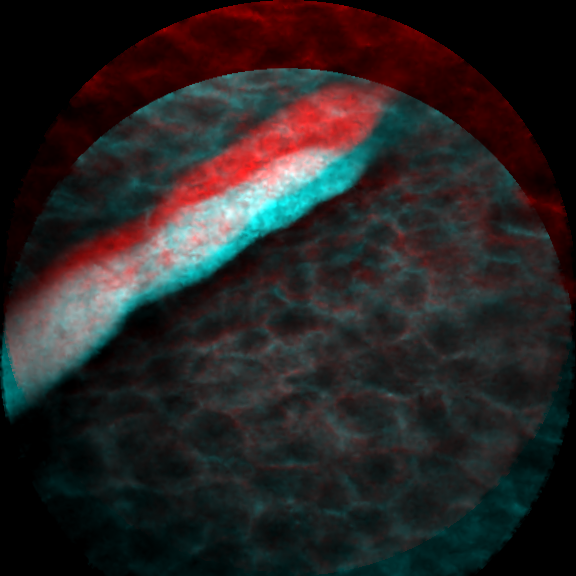}%
  {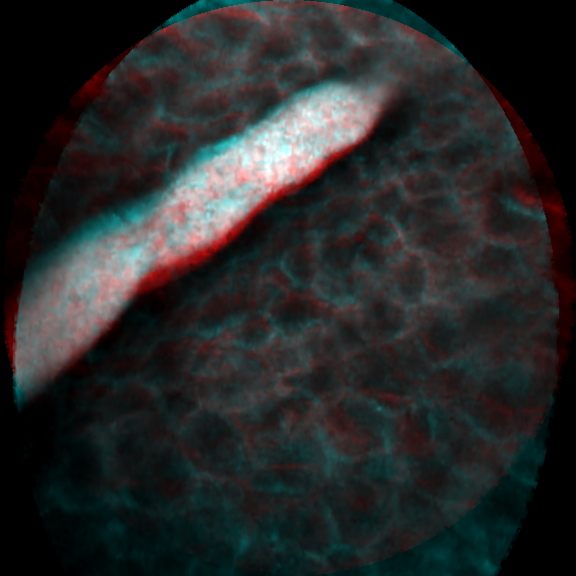}%
  {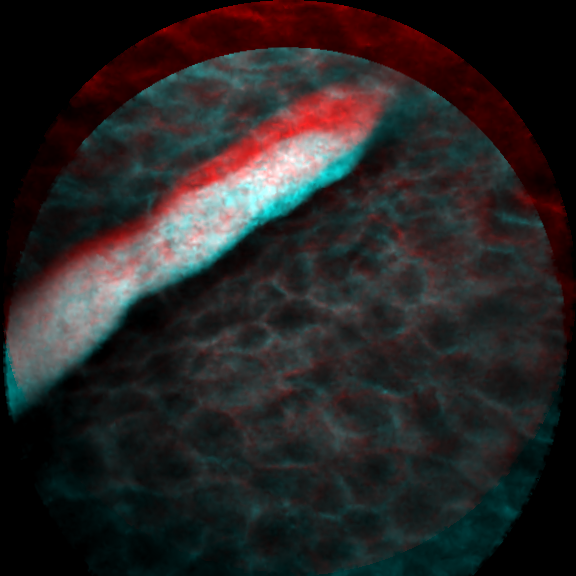}%
  {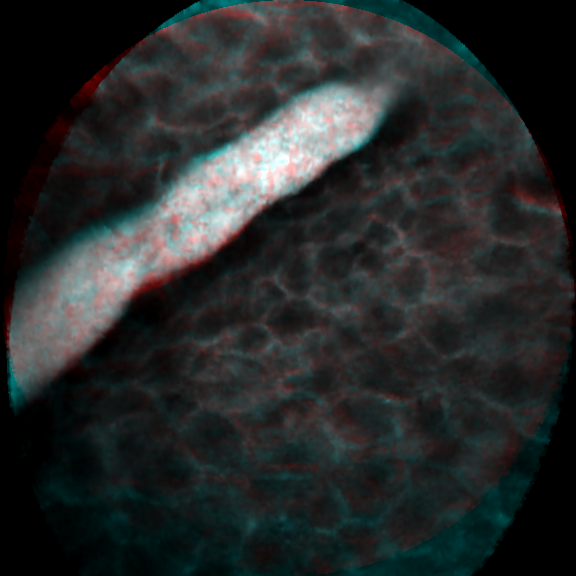}%
  {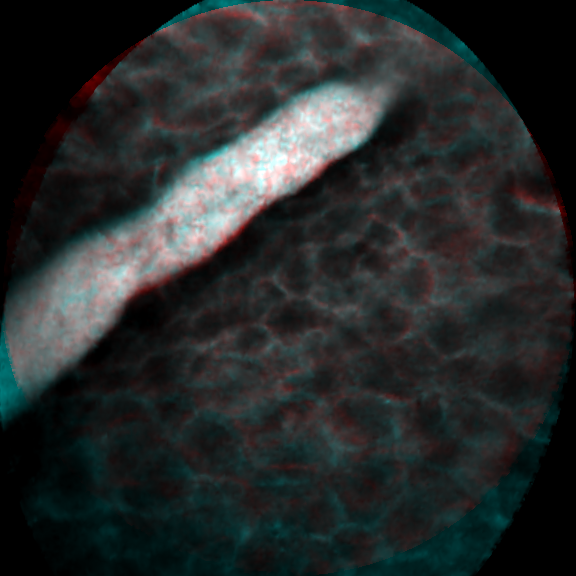}

\RegRow{182}{8--9}%
  {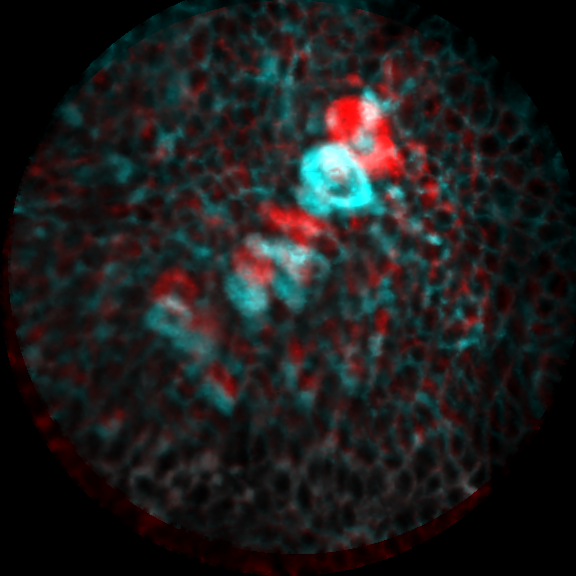}%
  {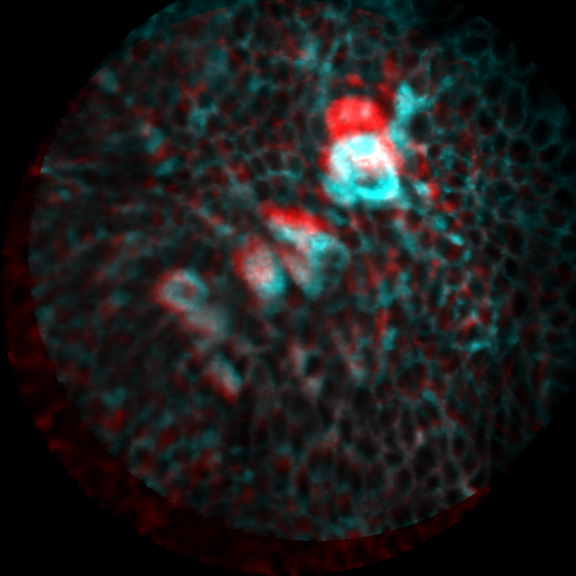}%
  {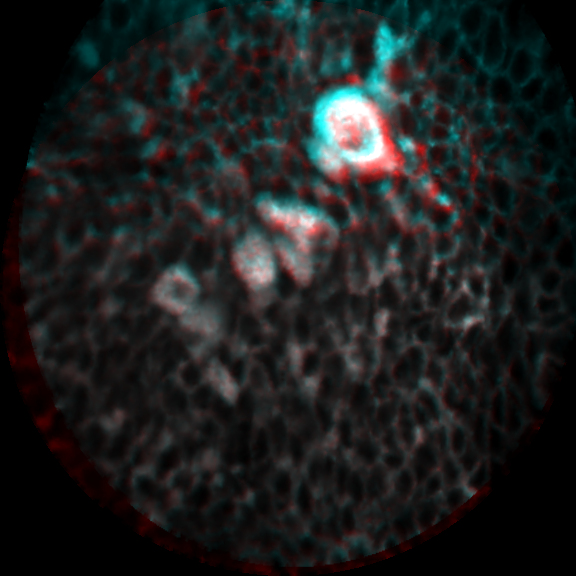}%
  {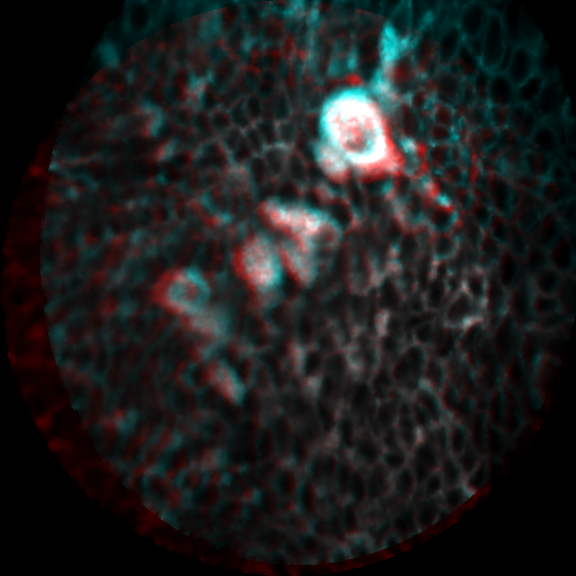}%
  {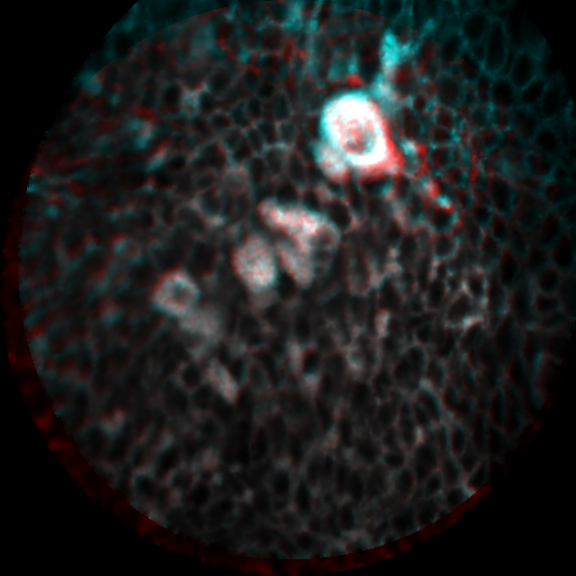}%
  {Optimal-Design-layout/images/landmark_registeration_182_pair_8_9/tps_raw_overlay.png}
\end{tabular}
}
\caption{Exemplary registration overlays based on reference landmarks under increasing transformation flexibility: (a) translation, (b) rigid, (c) affine, (d) TPS (Affine+RANSAC), (e) TPS (Raw), (f) TPS (Random Sampling). The leftmost column gives the sequence number and annotated frame-pair index. The top row illustrates limits of the constrained transformation models. While alignment in the bottom row is generally comparable for affine, TPS (Affine+RANSAC), and TPS (raw), remaining motion artifacts caused by non-continuous displacements are visible in the top part of the image. Additional visual examples can be found in the Suppl. Material.}
\label{fig:examples_landmark_overlays}
\end{figure}

\subsection{Alignment Based on Manual Landmarks}
\label{sec:results_landmark_based_registration}

Figure~\ref{fig:examples_landmark_overlays} provides qualitative examples of
the alignment obtained with the evaluated transformation families. 
Table~\ref{tab:landmark_results} summarizes the six transformation
configurations estimated from landmark correspondences. The transformation
configuration had a significant overall effect on both TRE (Friedman
$Q=51.347$, permutation $p=2.0\times10^{-5}$, Kendall's $W=0.734$) and NCC
($Q=55.878$, permutation $p=2.0\times10^{-5}$, $W=0.798$), indicating a strong
and consistent ordering across the 14 sequences. TPS with random sampling
obtained the lowest median TRE ($3.488$), the highest median NCC ($0.938$), and
the best mean rank for both metrics. Holm-adjusted post-hoc tests showed that it
outperformed every other configuration for both TRE and NCC. It was also
non-dominated in 13 of the 14 sequence-level Pareto comparisons and was the only
member of the aggregate empirical Pareto front. Translation and rigid
transformations were not significantly different from each other for either
metric after Holm correction.

\begin{table}[!t]
\centering
\caption{Sequence-level performance of landmark-derived transformation configurations.}
\label{tab:landmark_results}
\small
\resizebox{\textwidth}{!}{%
\begin{tabular}{lcccccc}
\toprule
Transformation configuration
& TRE median [IQR] $\downarrow$
& TRE rank $\downarrow$
& NCC median [IQR] $\uparrow$
& NCC rank $\downarrow$
& Pareto count
& Pareto rate \\
\midrule
Translation
& $7.895\;[5.182,10.780]$
& $5.07$
& $0.859\;[0.804,0.907]$
& $5.21$
& $0/14$
& $0.0\%$ \\

Rigid
& $6.986\;[4.836,10.633]$
& $5.14$
& $0.860\;[0.785,0.910]$
& $5.36$
& $0/14$
& $0.0\%$ \\

Affine
& $5.388\;[3.480,7.324]$
& $3.29$
& $0.899\;[0.836,0.928]$
& $3.21$
& $1/14$
& $7.1\%$ \\

TPS (Affine+RANSAC)
& $5.024\;[4.232,7.363]$
& $4.14$
& $0.894\;[0.838,0.921]$
& $3.86$
& $0/14$
& $0.0\%$ \\

\textbf{TPS (random sampling)}
& $\mathbf{3.488\;[3.254,4.008]}$
& $\mathbf{1.14}$
& $\mathbf{0.938\;[0.882,0.952]}$
& $\mathbf{1.07}$
& $\mathbf{13/14}$
& $\mathbf{92.9\%}$ \\

TPS (raw)
& $3.938\;[3.379,4.428]$
& $2.21$
& $0.930\;[0.883,0.950]$
& $2.29$
& $1/14$
& $7.1\%$ \\
\bottomrule
\end{tabular}}

\vspace{2pt}
\begin{minipage}{0.98\textwidth}
\footnotesize
\textit{Note:} Values are the median and interquartile range across 14
sequence-level summaries; rank~1 is best. The Pareto rate is the proportion of
sequences on which the configuration was non-dominated with respect to TRE and
$1-\mathrm{NCC}$. One landmark-TPS TRE value was missing; the corresponding
frame pair was excluded for every landmark configuration in the affected
sequence, so all configurations were compared using the same frame pairs.
\end{minipage}
\end{table}

\begin{figure}[t]
\centering
\includegraphics[width=\linewidth]{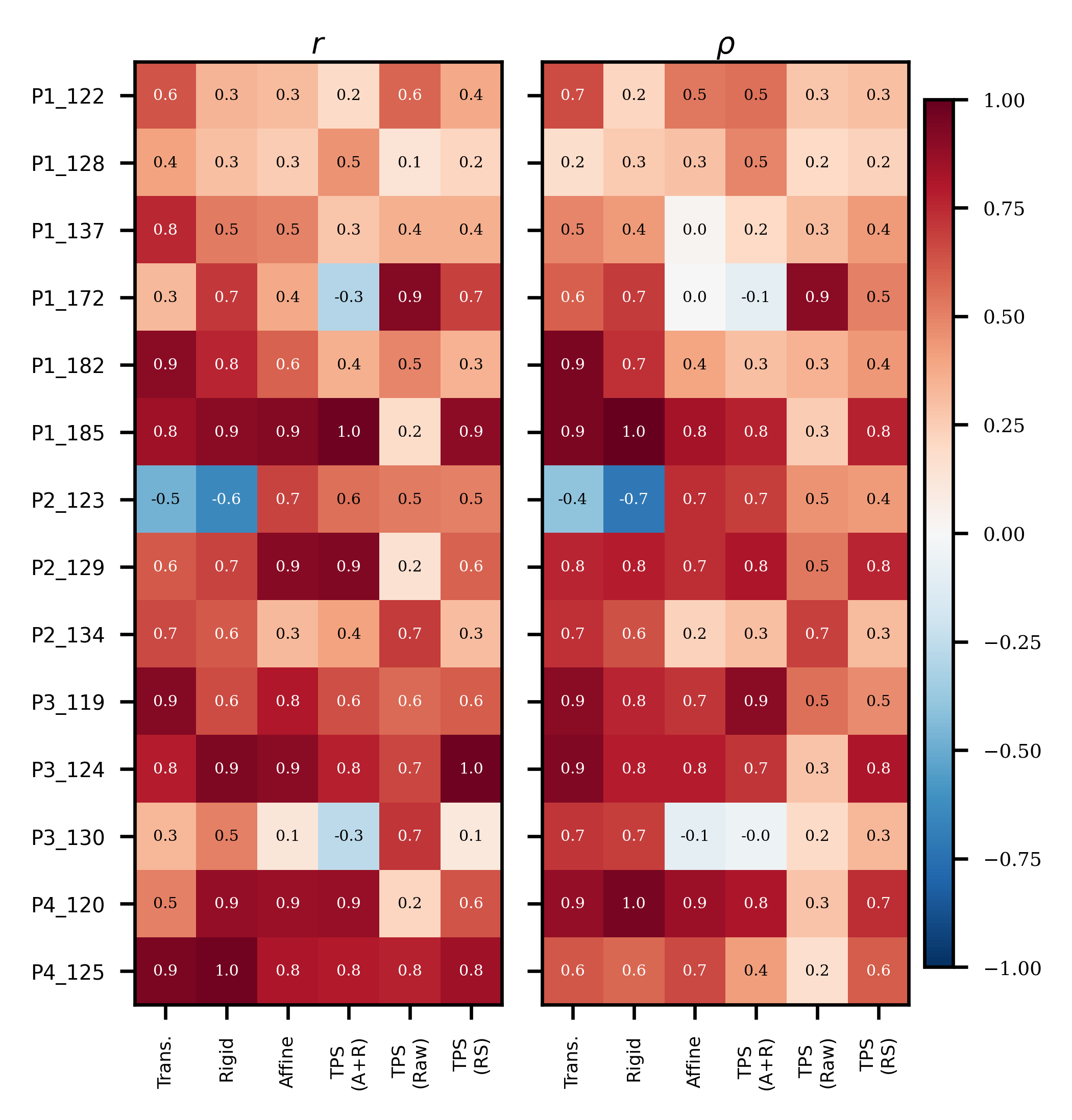}
\caption{Pearson's $r$ (left) and
Spearman rank correlation coefficient $\rho$ (right) between per-pair target
registration error (TRE, $L_2$) and photometric dissimilarity
($1-\mathrm{NCC}$), computed within each sequence across its annotated frame
pairs for each transformation configuration. All values use transformations
estimated from manually annotated landmark correspondences.}
\label{fig:tre_ncc_correlation_heatmaps}
\end{figure}

The correlation analysis in Figure~\ref{fig:tre_ncc_correlation_heatmaps}
supports treating TRE and 1-NCC as complementary measures. If both metrics
described the same aspect of registration performance, a consistently strong
positive association between TRE and $1-\mathrm{NCC}$ would be expected.
Instead, the correlations varied substantially across sequences and
transformation configurations, while differences between Pearson and Spearman
coefficients showed that the relationship was not consistently linear or
monotonic. This is consistent with the distinct quantities measured by the two
metrics: TRE evaluates geometric correspondence at the landmarks, whereas NCC
evaluates intensity-based image similarity. Their joint consideration
therefore provides a more complete evaluation of registration quality.

\subsection{Automatic Alignment Comparison}
\label{sec:methods-transformation-models-results}

% \begin{figure*}[!t]
% \centering
% \includegraphics[width=0.8\linewidth,keepaspectratio]{tre-boxplot-all-methods-cut}
% \caption{\textcolor{blue}{Per-pair TRE for all sequences and feature-matching
% backends under the affine and three TPS transformation configurations. The
% black dashed line shows the manual-landmark reference using TPS
% (Affine+RANSAC). Values above 40 pixels are omitted for clarity.}}
% \label{fig:tre-boxplot-all-methods-cut}
% \end{figure*}

Backend performance varied with the transformation configuration (Table~\ref{tab:backend_by_transform}). For translation, rigid, and affine registration, pairwise comparisons provided limited evidence of differences between backends after Holm correction. Across all three TPS configurations, RoMa consistently achieved the best mean rank for both TRE and NCC and was non-dominated in at least 13 of 14 sequences, with moderate agreement in backend rankings across sequences ($W = 0.38$--$0.51$).

\begin{table}[!t]
\centering
\caption{Best-ranked feature backend within each transformation configuration.}
\label{tab:backend_by_transform}
\scriptsize
\resizebox{\textwidth}{!}{%
\begin{tabular}{llccccccclc}
\toprule
Transformation configuration
& TRE leader
& TRE median [IQR]
& $p_{\mathrm{TRE}}$
& $W_{\mathrm{TRE}}$
& NCC leader
& NCC median [IQR]
& $p_{\mathrm{NCC}}$
& $W_{\mathrm{NCC}}$
& Pareto-rate leader
& Rate \\
\midrule
Translation
& SuperPoint+LK
& $7.321\;[4.654,10.373]$
& $0.0723$
& $0.142$
& SuperPoint+LK
& $0.848\;[0.793,0.920]$
& $0.0650$
& $0.146$
& SuperPoint+LK
& $7/14$ \\

Rigid
& SuperPoint+LK
& $7.210\;[4.294,9.703]$
& $0.2384$
& $0.097$
& RoMa
& $0.852\;[0.812,0.929]$
& $0.0399$
& $0.163$
& RoMa
& $6/14$ \\

Affine
& RoMa
& $3.342\;[2.827,5.076]$
& $0.0923$
& $0.133$
& RoMa
& $0.907\;[0.857,0.957]$
& $0.00730$
& $0.217$
& RoMa
& $13/14$ \\

TPS (Affine+RANSAC)
& RoMa
& $2.462\;[2.112,2.877]$
& $4.0\!\times\!10^{-5}$
& $0.437$
& RoMa
& $0.942\;[0.889,0.967]$
& $2.0\!\times\!10^{-5}$
& $0.514$
& RoMa
& $13/14$ \\

TPS (random sampling)
& RoMa
& $2.119\;[1.861,2.409]$
& $2.0\!\times\!10^{-5}$
& $0.406$
& RoMa
& $0.965\;[0.938,0.974]$
& $2.0\!\times\!10^{-5}$
& $0.492$
& RoMa
& $14/14$ \\

TPS (raw)
& RoMa
& $2.062\;[1.874,2.419]$
& $6.0\!\times\!10^{-5}$
& $0.377$
& RoMa
& $0.963\;[0.937,0.973]$
& $4.0\!\times\!10^{-5}$
& $0.407$
& RoMa
& $13/14$ \\
\bottomrule
\end{tabular}}

\vspace{2pt}
\begin{minipage}{0.98\textwidth}
\footnotesize
% \textit{Note:} The TRE and NCC leaders are selected independently according to
% the best mean within-sequence rank. Reported $p$-values and Kendall's $W$ refer
% to the six-backend Friedman comparison for the corresponding metric and
% transformation configuration, not to an individual pairwise contrast. A
% significant omnibus test does not imply that every pairwise contrast is
% significant. The Pareto-rate leader is the backend most frequently
% non-dominated within that transformation configuration.
\end{minipage}
\end{table}

In the transformation-averaged analysis, in which each configuration
received equal weight within each sequence (Table~4), the backend effect
remained significant for TRE and NCC (permutation
$p \leq 1.2 \times 10^{-4}$; $W = 0.32$ and $0.39$), and RoMa ranked
first for both metrics, was significantly better than every other
backend after Holm correction, and was non-dominated in all 14
sequences. This indicates robustness across the evaluated transformation
set rather than a transformation-independent backend effect.

\begin{table}[!t]
\centering
\caption{Transformation-averaged feature-backend robustness.}
\label{tab:backend_robustness}
\small
\resizebox{\textwidth}{!}{%
\begin{tabular}{lcccccc}
\toprule
Backend
& TRE median [IQR] $\downarrow$
& TRE rank $\downarrow$
& NCC median [IQR] $\uparrow$
& NCC rank $\downarrow$
& Pareto count
& Pareto rate \\
\midrule
LightGlue
& $4.998\;[3.923,5.880]$
& $3.93$
& $0.899\;[0.844,0.935]$
& $3.79$
& $1/14$
& $7.1\%$ \\

LoFTR
& $4.746\;[3.809,6.546]$
& $3.71$
& $0.893\;[0.847,0.913]$
& $4.29$
& $2/14$
& $14.3\%$ \\

\textbf{RoMa}
& $\mathbf{4.225\;[3.165,5.648]}$
& $\mathbf{1.36}$
& $\mathbf{0.909\;[0.868,0.951]}$
& $\mathbf{1.21}$
& $14/14$
& $100\%$ \\

Shi--Tomasi+LK
& $5.473\;[3.484,7.267]$
& $4.00$
& $0.882\;[0.857,0.944]$
& $4.00$
& $1/14$
& $7.1\%$ \\

SuperGlue
& $5.388\;[3.999,5.984]$
& $4.21$
& $0.894\;[0.843,0.930]$
& $4.29$
& $2/14$
& $14.3\%$ \\

SuperPoint+LK
& $4.931\;[3.514,7.575]$
& $3.79$
& $0.886\;[0.859,0.942]$
& $3.43$
& $1/14$
& $7.1\%$ \\
\bottomrule
\end{tabular}}

\vspace{2pt}
\begin{minipage}{0.98\textwidth}
\footnotesize
% \textit{Note:} For every sequence and backend, the six
% transformation-specific sequence summaries were averaged with equal weight
% before comparing backends. Sequence--transformation combinations were not
% treated as independent samples.
\end{minipage}
\end{table}

When each backend--transformation pair was treated as a complete
pipeline, the 36 pipelines differed significantly in overall ranking
(permutation $p = 2.0 \times 10^{-5}$ for both metrics), but none of the
630 pairwise contrasts per metric survived Holm correction, so no
individual pipeline can be declared statistically superior. The aggregate empirical Pareto front contained only RoMa with
TPS (raw), which had the lower median TRE, and RoMa with TPS (random
sampling), which had the higher median NCC and more frequent
sequence-level Pareto membership (13/14 versus 7/14); these Pareto
results are descriptive and comparison-set dependent (Suppl.
Mat., Sec.~3.4, Tab.~2).

\subsection{Sequence-Level Mosaicing}
\label{sec:sequence_level_mosaicing_results}

Sequence mosaicing was attempted for all combinations of 14
sequences, six automatic backends plus the landmark-guided reference, and
six transformation configurations (588 runs; completion and fragmentation
statistics in Suppl. Mat., Sec.~3.5.1, Tab.~3). Translation,
rigid, affine, and TPS (Affine+RANSAC) completed essentially on all runs,
whereas the less-filtered TPS variants failed on a minority of
automatic-correspondence runs, and on none of the landmark-guided ones,
because of implausibly large predicted frame bounds, indicating greater
sensitivity to poorly distributed or ill-conditioned correspondences.

\subsubsection{TPS random sampling backend comparison}

\begin{figure}[t]
  \centering
  \begin{minipage}{0.8\linewidth }
  \centering
    \includegraphics[width=0.9\linewidth]{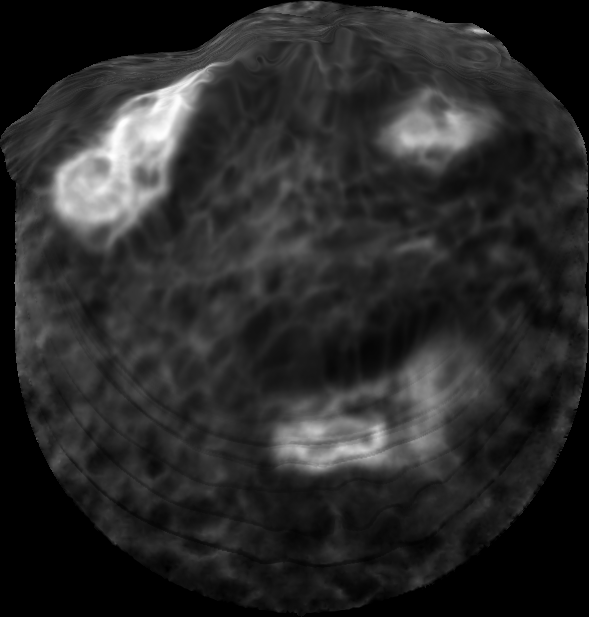}
  \end{minipage}\hfill
  \caption{Example pCLE mosaic from sequence 125, reconstructed using RoMa feature matching and TPS with random sampling.}
  \label{fig:mosaic}
\end{figure}

Because the pairwise analysis identified TPS with random sampling as
the strongest transformation configuration (Sec.~4.2), the sequence-level
backend comparison focuses on it, to determine whether its pairwise
advantage persists after sequential transformation accumulation; this
selection rests on the pairwise analysis, not on the mosaic results.
Table~\ref{tab:mosaic_backend_comparison} summarizes final-mosaic quality
across the completed mosaics. RoMa achieved the highest median LOFO-NCC
and the lowest median mosaic-coordinate TRE but completed 12 of 14
sequences, whereas LightGlue completed all 14 and achieved the lowest
median DRA among the automatic methods, a trade-off between conditional
quality and execution reliability. Restricting the comparison to the 12
sequences reconstructed by both RoMa and LightGlue did not change this
ordering (Suppl. Material, Sec.~3.5.2, Table~4), so RoMa's
conditional quality advantage is not explained by the different
evaluated sequence sets. The two sequences RoMa did not reconstruct
(CLE\_VF\_P1\_128 and CLE\_VF\_P2\_123) were challenging for the
alternatives as well (Suppl. Material, Sec.~3.5.2, Table~5); on
CLE\_VF\_P1\_128, LightGlue reached a LOFO-NCC of 0.939 while its
mosaic-coordinate TRE was 31.45\,px, illustrating that high
leave-one-frame-out photometric consistency does not imply geometrically
consistent mosaicing. Figure~\ref{fig:mosaic} shows the dataset’s longest consecutive sequence, with peripheral distortions potentially arising from TPS extrapolation beyond matched regions. As correspondences are confined to the overlap, constraining deformation outside their spatial support remains an open challenge for this pipeline. Additional qualitative results are shown in Supp. Mat. Figs. 10–12.

\begin{table}[t]
\centering
\small
\caption{Sequence-level backend comparison for TPS with random sampling,
reported as median [interquartile range] across the available successfully
completed mosaics. Higher LOFO-NCC is preferred, whereas lower
mosaic-coordinate TRE and DRA are preferred. Bold values identify the most
favorable automatic-backend result.}
\label{tab:mosaic_backend_comparison}
\resizebox{\textwidth}{!}{%
\begin{tabular}{lcccc}
\toprule
Registration source & Completed & LOFO-NCC $\uparrow$ &
Mosaic-coordinate TRE (px) $\downarrow$ & DRA (px) $\downarrow$ \\
\midrule
RoMa           & 12/14 & \textbf{0.919} [0.894, 0.945] & \textbf{2.88} [2.61, 4.06] & 5.66 [2.82, 7.67] \\
LightGlue      & 14/14 & 0.909 [0.883, 0.916] & 4.44 [3.50, 4.99] & \textbf{3.55} [2.39, 7.08] \\
SuperGlue      & 12/14 & 0.886 [0.868, 0.900] & 4.47 [3.78, 5.41] & 6.13 [3.88, 8.54] \\
Shi--Tomasi+LK & 13/14 & 0.868 [0.844, 0.905] & 15.21 [4.20, 20.77] & 29.78 [5.50, 42.71] \\
SuperPoint+LK  & 14/14 & 0.867 [0.832, 0.905] & 13.72 [5.42, 18.67] & 16.97 [9.19, 62.78] \\
LoFTR          &  7/14 & 0.827 [0.738, 0.863] & 43.74 [33.59, 56.06] & 139.13 [72.69, 165.56] \\
\midrule
Landmark-guided reference & 14/14 & 0.883 [0.846, 0.917] & 5.23 [4.40, 6.29] & 3.37 [1.81, 4.08] \\
\bottomrule
\end{tabular}}
\end{table}

\subsubsection{Pairwise-to-mosaic consistency}
\label{sec:pairwise_mosaic_consistency}

We finally examined whether pairwise registration quality is predictive
of final-mosaic quality, correlating each sequence's pairwise-level
metric against its corresponding mosaic-level metric across all
transformation configurations and matching backends (full statistical
detail in Suppl. Mat., Sec.~3.5). Table~\ref{tab:pairwise_to_mosaic}
(top) summarizes that across all tested configurations, the association
was weak overall and reached stat. significance only for a small
minority, concentrated on pairwise NCC predicting geometric
TRE.

% \begin{table}[t]
% \centering
% \setlength{\tabcolsep}{2pt}
% \caption{Association between pairwise and mosaic-level registration
% quality, pooled across all tested transformation configurations and
% matching backends.}
% \label{tab:pairwise_mosaic_general}
% \small
% \begin{tabular}{lccc}
% \toprule
% Predictor $\to$ Outcome & $n$ & Sig.\ (FDR) & Median $r$ \\
% \midrule
% Pairwise NCC $\to$ Mosaic landmark TRE & 42 & 8 & $-0.39$ \\
% Pairwise NCC $\to$ Mosaic DRA & 42 & 4 & $-0.23$ \\
% Pairwise TRE $\to$ Mosaic LOFO-NCC & 42 & 0 & $-0.19$ \\
% \bottomrule
% \end{tabular}
% \end{table}

Broken down by transformation configuration (Suppl. Mat., Sec.~3.5.3, Table~6), this strongest
relationship, pairwise NCC against mosaic landmark TRE, was concentrated
in the models already shown to be geometrically insufficient for this
task: 3 of 7 rigid and 2 of 7 translation configurations reached
significance, against 0 to 2 of 7 for affine and the three TPS variants.
We do not consider this informative: a model that performs poorly and
unevenly will trivially show pairwise-to-mosaic association; that
association says nothing about whether pairwise quality is a useful
predictor for a model one would actually deploy.

% \begin{table}[t]
% \centering
% \caption{Pairwise NCC $\to$ mosaic-coordinate landmark TRE, broken down
% by transformation configuration and pooled across matching backends.}
% \label{tab:pairwise_mosaic_by_model}
% \small
% \begin{tabular}{lcc}
% \toprule
% Transformation configuration & $n$ & Sig.\ (FDR) \\
% \midrule
% Translation & 7 & 2 \\
% Rigid & 7 & 3 \\
% Affine & 7 & 0 \\
% TPS (Affine+RANSAC) & 7 & 0 \\
% TPS (Raw) & 7 & 1 \\
% TPS (Random Sampling) & 7 & 2 \\
% \bottomrule
% \end{tabular}
% \end{table}

Restricting attention to the transformation configurations with adequate
representational capacity, affine and the three TPS variants, associations
became sparse rather than clearer. Narrowing further to RoMa and
LightGlue, the two backends favored in this evaluation for
accuracy and reliability (Sec.~4.3, Table~5), the association largely
disappeared (Table~\ref{tab:pairwise_to_mosaic}, bottom): almost none of
these configurations reached significance, and the typical correlation
across this subset was close to zero. This pattern is consistent with
restriction of range: these configurations already perform consistently
well across nearly all sequences, leaving little cross-sequence
variability for a correlation to detect, unlike the uneven performance of
translation and rigid registration above.

\begin{table}[t]
\centering
\setlength{\tabcolsep}{2pt}
\small
\caption{Association between pairwise and mosaic-level registration
quality. Top: pooled across all tested transformation configurations and
matching backends. Bottom: restricted to representationally adequate
transformation configurations (affine, TPS variants) and to RoMa and
LightGlue. $n$ is the number of backend--transformation combinations
tested; Sig.\ (FDR) is the number reaching significance after
Benjamini--Hochberg correction.}
\label{tab:pairwise_to_mosaic}
\begin{tabular}{lccc}
\toprule
Predictor $\rightarrow$ Outcome & $n$ & Sig.\ (FDR) & Median $r$ \\
\midrule
\multicolumn{4}{l}{\textit{All configurations and backends}} \\
Pairwise NCC $\rightarrow$ Mosaic landmark TRE & 42 & 8 & $-0.39$ \\
Pairwise NCC $\rightarrow$ Mosaic DRA          & 42 & 4 & $-0.23$ \\
Pairwise TRE $\rightarrow$ Mosaic LOFO-NCC     & 42 & 0 & $-0.19$ \\
\midrule
\multicolumn{4}{l}{\textit{Affine and TPS variants; RoMa and LightGlue}} \\
Pairwise NCC $\rightarrow$ Mosaic landmark TRE & 8 & 1 & $0.26$ \\
Pairwise NCC $\rightarrow$ Mosaic DRA          & 8 & 1 & $0.32$ \\
Pairwise TRE $\rightarrow$ Mosaic LOFO-NCC     & 8 & 0 & $0.16$ \\
\bottomrule
% \multicolumn{4}{l}{\footnotesize $^{\dagger}$Median of $|r|$.}
\end{tabular}
\end{table}

Taken together, these results indicate that pairwise registration
performance cannot reliably be used to anticipate final mosaic quality
under pairwise registration alone, and that this limitation is most
pronounced for exactly the configurations this evaluation would otherwise
recommend. Mosaic quality should therefore continue to be assessed
directly rather than inferred from pairwise-level metrics.

\section{Conclusion}

We characterize the landmark-derived reference performance of four
transformation models on real clinical pCLE frames and use it to
evaluate six feature-matching backends. Prior work has mostly assessed
mosaicing quality with the photometric criterion optimized during
registration or under laboratory conditions; with the exception
of~\cite{Hao:26}, landmark-based validation on clinical pCLE data has
been absent. Translation and rigid models proved insufficient under
tissue deformation, and TPS with random sampling achieved the best
landmark-derived TRE and NCC. Among the backends, only RoMa offered a
consistent advantage: it ranked first for both metrics in the
transformation-averaged comparison and was significantly better than
every other backend after Holm correction, whereas backend effects were
weak for translation, rigid, and affine registration. At the sequence
level, RoMa produced the highest-quality mosaics while LightGlue
completed every sequence with less directional error, and pairwise
registration quality was a weak predictor of mosaic quality. TRE and NCC
should be treated as complementary rather than redundant. Future work
will add CWCR~\cite{Gong2021RobustMosaicing} and TESCV~\cite{Gong2022IntensityBased}, and the
motion-aware deformation model of~\cite{Hao:26}, since probe motion can
introduce discontinuities that the models evaluated here cannot
represent; further steps are fine-tuning the learned matchers on this
dataset and expanding the annotations.

% In this paper, we investigate ... establish a baseline for the six annotated sequences as to the best performance that can be expected from each transformation model for each pair in each sequence. 
% %This was necessary since obtaining an NCC of 1.0 is implausible due to the presence of noise. 
% Our results clearly demonstrate that translation-only and rigid transformations result in considerable residual alignment errors for CLE mosaicing, as they generate large errors once the tissue deforms. A still-open question that needs to be investigated is whether Affine will always be sufficient for pCLE images regardless of the degree of deformation, or whether TPS outperforms it under specific conditions; designing an experiment to test this empirically may be challenging, given the difficulty of annotating landmarks on extremely deformed tissue. Another open question is how to fine-tune the learning-based feature detection and extraction methods to obtain better features, and whether this is even necessary, since Shi-Tomasi + Lucas-Kanade consistently performs comparably to the learning-based methods. VoxelMorph \cite{balakrishnan2019tmi} is another approach worth considering in future work, as it is an unsupervised method that could bypass the problem of acquiring annotated datasets we currently face.

% \bmsubsection*{Acknowledgments}
% TODO

% \bmsubsection*{Financial Disclosure}
% TODO

% \bmsubsection*{Conflicts of Interest}

% TODO

\FloatBarrier
\bibliographystyle{splncs04}  % or whichever style your venue requires
\bibliography{wileyNJD-Chicago}  
% \nocite{*}% Show all bib entries - both cited and uncited; comment this line to view only cited bib entries;
\end{document}